\documentclass{article}

\usepackage{PRIMEarxiv}

\usepackage[utf8]{inputenc} 
\usepackage[T1]{fontenc}    
\usepackage{hyperref}       
\usepackage{url}            
\usepackage{booktabs}       
\usepackage{amsfonts}       
\usepackage{nicefrac}       
\usepackage{microtype}      
\usepackage{lipsum}
\usepackage{fancyhdr}       
\usepackage{graphicx}       
\usepackage{xcolor}
\usepackage{todonotes}
\graphicspath{{media/}}     

\usepackage[normalem]{ulem} 

\usepackage{listings}   
\definecolor{codegreen}{rgb}{0,0.6,0}
\definecolor{codegray}{rgb}{0.5,0.5,0.5}
\definecolor{codepurple}{rgb}{0.58,0,0.82}
\definecolor{backcolour}{rgb}{0.95,0.95,0.92}
\definecolor{lightgray}{rgb}{.9,.9,.9}
\definecolor{darkgray}{rgb}{.4,.4,.4}
\definecolor{purple}{rgb}{0.65, 0.12, 0.82}

\lstdefinelanguage{JavaScript}{
  keywords={typeof, new, true, false, catch, function, return, null, catch, switch, var, if, in, while, do, else, case, break},
  keywordstyle=\color{blue}\bfseries,
  ndkeywords={class, export, boolean, throw, implements, import, this},
  ndkeywordstyle=\color{darkgray}\bfseries,
  identifierstyle=\color{black},
  sensitive=false,
  comment=[l]{//},
  morecomment=[s]{/*}{*/},
  commentstyle=\color{purple}\ttfamily,
  stringstyle=\color{red}\ttfamily,
  morestring=[b]',
  morestring=[b]"
}

\title{GrADyS-GS - A ground station for managing field experiments with Autonomous Vehicles and Wireless Sensor Networks}

\author{
  Breno Perricone, Thiago Lamenza, Marcelo Paulon, Bruno José Olivieri de Souza,  Markus Endler  \\
  Laboratory for Advanced Collaboration, Departmento de Informática \\
  Pontificial Catholic University of Rio de Janeiro (PUC-Rio)\\
  Rio de Janeiro, Brazil\\
}

\begin{document}
\maketitle

\begin{abstract}

In many kinds of research, collecting data is tailored to individual research. It is usual to use dedicated and not reusable software to collect data. GrADyS Ground Station framework (GrADyS-GS)  aims to collect data in a reusable manner with dynamic background tools. This technical report describes GrADyS-GS, a ground station software designed to connect with various technologies to control, monitor, and store results of Mobile Internet of Things field experiments with Autonomous Vehicles (UAV) and Sensor Networks (WSN). In the GrADyS project GrADyS-GS is used with ESP32-based IoT devices on the ground and Unmanned Aerial Vehicles (quad-copters) in the air. The GrADyS-GS tool was created to support the design, development and testing of simulated movement coordination algorithms for the AVs, testing of customized Bluetooth Mesh variations, and overall communication, coordination, and context-awareness field experiments
planed in the GraDyS project. Nevertheless, GrADyS-GS is also a general purpose tool, as it relies on a dynamic and easy-to-use Python and JavaScript framework that allows easy customization and (re)utilization in another projects and field experiments with other kinds of IoT devices, other WSN types and protocols, and other kinds of mobile connected flying or ground vehicles. So far, GrADyS-GS has been used to start UAV flights and collects its data in s centralized manner inside GrADyS project.

\end{abstract}


\section{Introduction}

While developing new mobile communication solutions, simulation is a powerful tool for testing, analyzing the performance, and assessing the scalability of the protocols. 
Nevertheless, when it comes to the deployment of the research-based and  engineered system in the wild, and subject to tests in field experiments, then many new challenges  come into play. 
Unlike simulations, field tests usually reveal many unforeseen and unexpected  behaviors, related to various causes, and which add up to the long list of "external" factors to consider. This all turns the field experiments much more unpredictable than simulated experiments, and more often than not, the operator of the field test operator must thoroughly analyze the behaviour and the data exchanges of the (sometimes mobile) system elements after the test is concluded. 
This calls for a tool to control and drive the field experiments, visualize and save experiment-related data for later debugging, and hence perform these tests efficiently.

This work is part of the results generated by the research entitled GrADyS \cite{gradys2021} (Ground-and-Air Dynamic sensors networkS). The GrADyS\footnote{http://www.lac.inf.puc-rio.br/index.php/gradys/} project was created to investigate the applications of these networks in the monitoring of remote, dangerous, or hard-to-reach regions through the collection of sensor data using UAV swarms.

In the GrADyS\cite{gradys2021} (Ground-and-Air Dynamic sensors networkS) project, there are several cooperating components and protocols  in the field such as: 
(a) the flight trajectory of the UAVs is controlled through  ArduPilot\footnote{https://ardupilot.org/}; 
(b) a distributed algorithm, such as such as DADCA \cite{bolivieri2020}, coordinates the movements of the UAV ensamble,  relying on single board computers as RaspBerry Pi\footnote{https://www.raspberrypi.org/} aboard the UAVs; 
(c) Wifi and BlueTooth radios with custom firmware running in ESP32-based\footnote{https://www.espressif.com/en/products/socs/esp32} devices. Each of these technologies has its development, control, and monitoring stack, making the field experiments to be quite complex.

In many kinds of research, collecting data is tailored to individual research. It is usual to use dedicated and not reusable software to collect data. GrADyS Ground Station framework (GrADyS-GS)  aims to collect data in a reusable manner with dynamic background tools.

This technical report describes the GrADyS-GS, created to support the project's field testing. The ground station is a web application to monitor, control, and display device networks while storing the experiments' data for later analysis. 

The GrADyS-GS contains two primary sub-modules, with the Python programming language and Django Framework, and the front-end module, built with HTML, CSS, and Javascript programming language. Both modules communicate via WebSocket connections.

Currently, GrADyS-GS can connect to and control swarms of UAVs and WSN by organizing and facilitating field experiments that may have been  previously simulated using GrADyS-SIM\cite{gradyssim2022}.


\section{Django/Javascript}

The project's back-end relies on the Django framework, a tool written in Python, for fast and easy web development. It provides an easy-to-setup developing server, an extensible simple URL route map, and a highly modular architecture.
The interface logic relies on the JavaScript programming language, the most popular programming language in use.

Both technologies were chosen due to their wide use and popularity, providing a framework with high extensible usability. This ground station has a flexible foundation while built inside the GrADyS project's requisites. It provides an easy to extend functionality for command buttons, modular components to accept different communication protocols, and highly parameterized components accessible from configuration files.

\section{Architecture}

The GrADyS-GS is structured following the classic web development concept, with the Front-end module responsible for the interface and visualization and Back-end module responsible for server-side information processing and communication with external devices. Front-end is built with Javascript language, HTML, or Template language from Django, and Cascading Style Sheets (CSS) language. The back-end is mainly built with Python language, using Django Framework. Both modules communicate with each other via WebSocket channels. A socket connection is a dedicated full-duplex channel based on the Transmission Control Protocol (TCP). This project uses Django Channels library to handle WebSockets communication.

\begin{figure}[h]
    \includegraphics[width=1\textwidth, height=0.5\textwidth]{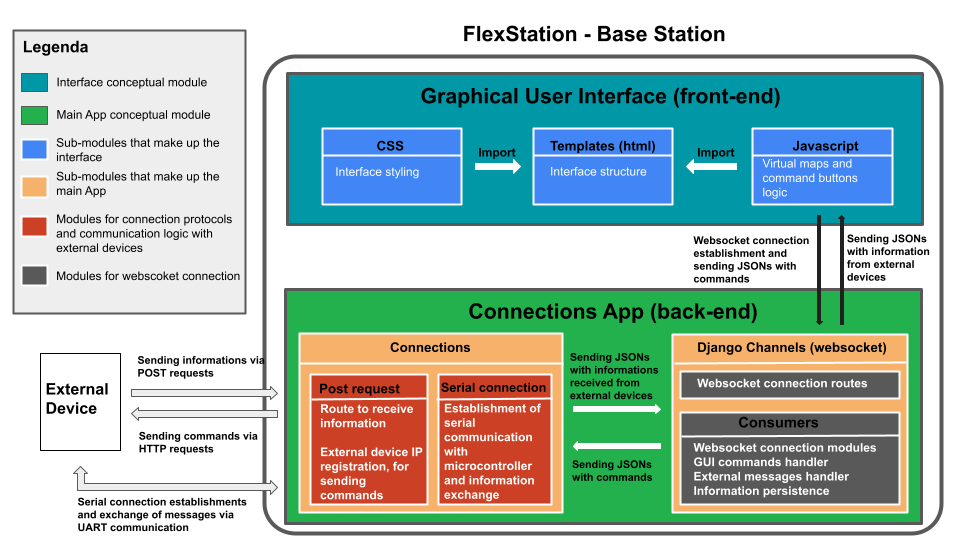}
    \caption{Project's Architecture}
\end{figure}

\subsection{Back-end}

Understanding the server-side structure of this project first required a basic understanding of how Django is structured and how it operates.

\subsubsection{Server}

Django provides a lightweight development Web server that can be used via the \verb|manage.py| file. By default, the server runs on port 8000 on the IP address 127.0.0.1 and should not be used on production. It's runnable with the command line:

\begin{lstlisting}[language=Python, caption= Starting server on Windows]
Windows
C:\path-to-this-cloned-repository\> python manage.py runserver
\end{lstlisting}

\begin{lstlisting}[language=Python, caption= Starting server on Linux]
Linux
gradys-gs$ python manage.py runserver
\end{lstlisting}

With the server up, when the browser accesses the home page, Django will call the corresponding python method, explained in the next section. In this project, the python method called when the home page is accessed will render the index HTML file.

\subsubsection{URL/View}
\label{architecture_backend_url_view}

Building a URL scheme with Django is a simple task, thanks to the URL/View mapping that the python web framework provides. When a user requests a page from the URL schema, Django does mapping to the corresponding Python function named View.
So, for example, the URL scheme below has a mapping between the \textbf{home page} path and \textbf{index} view, also between \textbf{/command/} path (note that 'command' is a simple integer) and \textbf{receive\_command\_test} view.

\begin{lstlisting}[language=Python, caption= URL view mapping example]
urlpatterns = [
    path('', index), 
    path('<int:command>/', receive_command_test),
]
\end{lstlisting}

Inside the main app's folder, \textit{connections}, there is \verb|urls.py| and \verb|views.py| files. The \verb|urls.py| file is responsible for associating a URL address and a view. Note that there is another \verb|urls.py| file inside the \textit{config} folder that is responsible for the whole project's pathing. So, for example, if there was another app in this project, a prefix path could be created to that specific app. The main app has the default path, so there is no prefix attached.

If it is needed to add a new URL path, it should be added a new \textbf{path()} item inside the URL patterns list, in \verb|/connections/urls.py|. For example:

\begin{lstlisting}[language=Python, caption=/connections/urls.py]
urlpatterns = [
    path('', index),
    path('<int:command>/', receive_command_test),
    path('new-path/', new_view)
]
\end{lstlisting}

Now it is necessary a view to handle the new URL path request. A \textit{view} is a Python function that takes a Web request and returns a Web response. This response can be the HTML contents of a Web page, or a JSON or a redirect, or a 404 error, or anything. The view itself contains whatever arbitrary logic is necessary to return that response.

\begin{lstlisting}[language=Python, caption=/connections/views.py, label={lst:label}]
def index(request):
  context = {
    'google_maps_key': settings.GOOGLE_MAPS_API_KEY
  }
  return render(request, 'index.html', context=context)
\end{lstlisting}
The Listing \ref{lst:label} represents the index view, accessed when the home page is loaded. It receives a request, creates a context variable with the google maps key from .env, and load the \verb|index.html| template, attached with the context. We store our views inside \verb|/connections/views.py|. 

To send additional parameters, it can be sent via the url body, for example, localhost:8000/new-path/10/. This URL need to be declared inside the \verb|connections/urls.py| as:

\begin{lstlisting}[language=Python, caption=Declaring URL with parameter]
path('new-path/<int:id>/', new_view)
\end{lstlisting}

Furthermore, the new view can receive an id parameter, as follows:

\begin{lstlisting}[language=Python, caption=View receiving URL parameter]
def new_view(request, id):
  # Function Logic
  return 
\end{lstlisting}

Now, accessing the default server 127.0.0.1:8000/new-path/5/will call the new\_view method, sending the parameter 5.

This URL pathing, provided by Django's framework, enables external devices to send POST Requests with information on the message's body. The corresponding View is in charge of the logic on how to handle the message. It can save the data received, handle errors, return an acknowledge message, register the device on a session list, forward to the front-end the JSON with the information received, and anything the application need for that URL.

\subsection{Front-end}

The front-end consists of templates files, CSS styling files, and Javascript files. As shown above, the home page template file is rendered when the default ip+port is accessed. New templates files can be added inside the /templates/ folder. They work very similarly to HTML files, with some add-ons.

\begin{lstlisting}[language=Python, caption=Template tag example]
{% load static %}
<link rel="stylesheet"  href="{% static 'connections/css/connection.css' %}">
\end{lstlisting}

The code above introduces the '\{\% \%\}' tag (that is not HTML native), in this case, to load a CSS file to the page.

Load a Javascript file in a template, the logic is the same, as long this Javascript file is inside the folder that STATIC\_URL variable is pointing to. This variable is inside config/settings.py. In our case, the STATIC\_URL variable is pointing to /static/ folder.

\begin{lstlisting}[language=Python, caption=config/settings.py]
STATIC_URL = '/static/'
\end{lstlisting}

The \verb|index.html| home page template loads \verb|gmap.js|, responsible for Google Map's virtual map and \verb|main.js|, which is responsible for starting websockets connections with the back-end and for the button's logic.

\subsection{Communication}

\begin{figure}
    \centering
    \includegraphics[width=1\textwidth, height=1\textwidth]{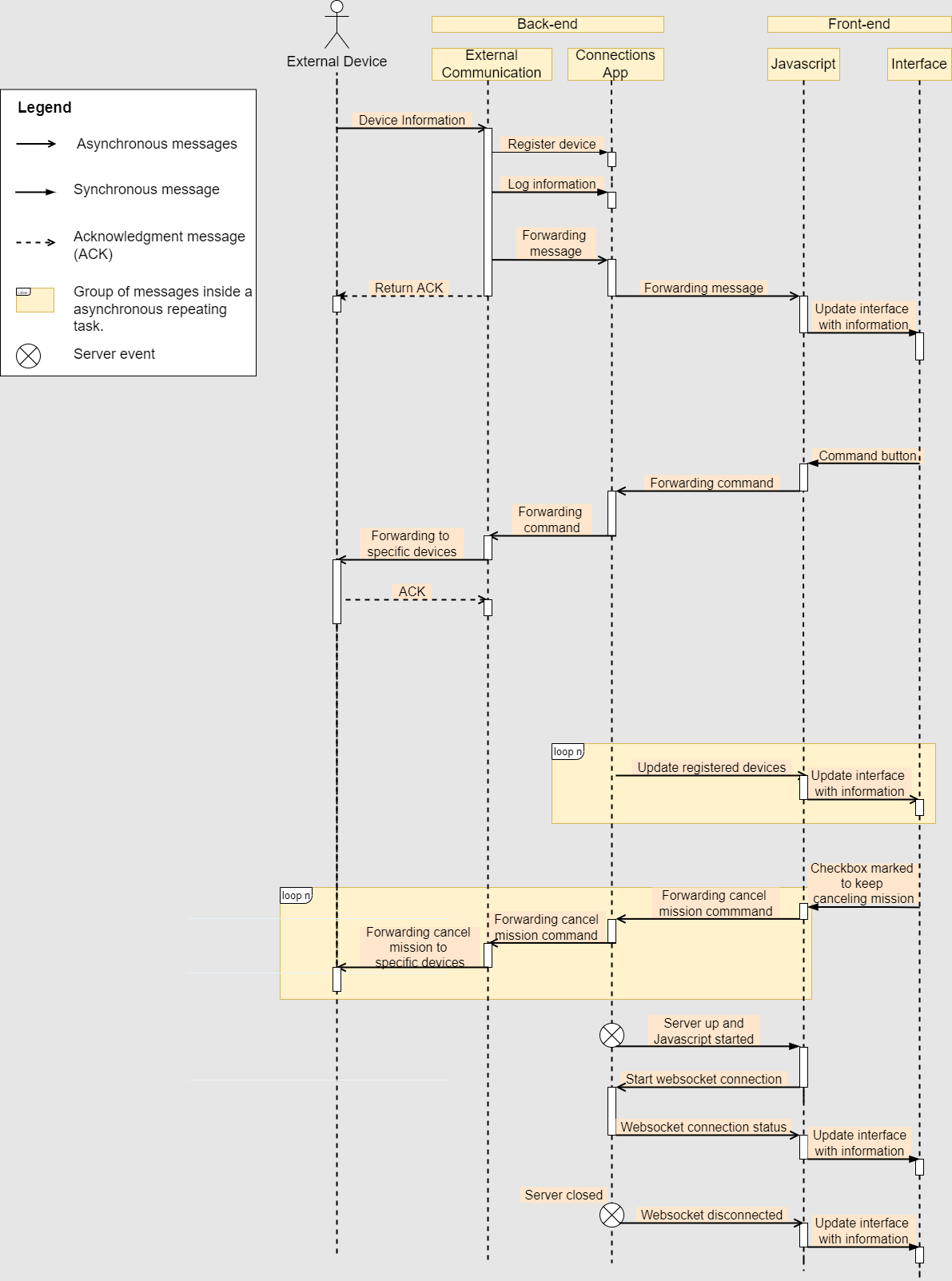}
    \label{fig:message}
    \caption{Event Diagram for the project's most common interactions}
\end{figure}

The sequence message diagram, Figure \ref{fig:message} represents the message flow between external devices and the main modules from this framework.

The two main modules communicate with each other, exchanging JSON messages and external devices with the back-end. 

The submodule, containing Javascript files, starts the socket connection with a route stated inside the Django Channels submodule. 
The information gate of the ground station to external devices is through the Connections submodule, which contains the routes and logic to receive/send information.

Note that the message protocol between the framework and external devices can differ from project to project, changing the way the information is delivered, or the commands are handled. But, the messages flow between the back-end and front-end modules should remain similar to this diagram.
Important things to notice are:

\begin{itemize}
    \item When a device sends information to the framework, the back-end will register this device. If not already on the persistent list, log the information and forward it to the front-end, to update the interface.
    \item There is a Consumer in charge to keep the persistent device list, with the registered devices, inside /connections/consumers\_wrapper/update\_periodically.py. In this Consumer, there is a task to update the activity status of the devices on the list, every X seconds, specified at config.ini. This is represented on the third group of messages flow in the diagram above.
    \item There is the possibility to create checkbox buttons that will trigger a constant task while the checkbox is pressed. This is represented in the fourth group of messages flow in the sequence diagram.
\end{itemize}

\subsubsection{Internal Communication}
The main app form of communication with templates, or HTML, is using WebSocket connections. Django Channels package mediates these connections. The logic to establish a WebSocket connection is similar to the URL/View logic presented on item \ref{architecture_backend_url_view}. The connections/routing.py file contains the WebSocket URL patterns, or schema:

\begin{lstlisting}[language=Python, caption=/connections/routing.py]
ws_urlpatterns = [
  path('ws/connection/', ConnectionConsumer.as_asgi()),
  path('ws/receive/', ReceiveCommandConsumer.as_asgi()),
]
\end{lstlisting}

As previously mentioned, Django Channels makes a mapping, associating an URL with a Consumer. A Consumer is a Python Class that handles a WebSocket connection. So, when our JavaScript is loaded, it tries to connect with a specific Consumer, accessing a specific URL inside the ws\_urlpatters.

\begin{lstlisting}[language=JavaScript, caption=Example of Websocket connection declaration]
// Javascript stablishing new connection
var socket = new WebSocket('ws://localhost:8000/ws/connection/');
\end{lstlisting}

When this command is received, the ConnectionConsumer class is called, and a connection is initiated. The Consumers are inside connections/consumers\_wrappers/ and a new one can be created, inheriting WebsocketConsumer or AsyncWebsocketConsumer, depending on its functionality. You can substitute three main methods:

\begin{itemize}
    \item connect: called when the specific url is accessed and start a dedicated connection with self.accept. This is the only method you NEED to override.
    \item receive: called when a message is sent via socket connection.
    \item disconnect: called when the connection is closed
\end{itemize}

Creating a new Consumer, is simple as creating a new file inside connections/consumers\_wrapper/ with a Class like:

\begin{lstlisting}[language=Python, caption=New Consumer example]
class NewConsumer(AsyncWebsocketConsumer):
  async def connect(self):
    await self.accept()

  async def receive(self, message):
    # Handle the message

  async def disconnect(self, close_code):
    # Handle disconnection

  async def additional_method(self, *args):
    # Additional method's logic
\end{lstlisting}

To send a message to the other side of connection (Django -> Javascript) it can be done using the send method, inherted from WebSocket class:

\begin{lstlisting}[language=Python, caption=Sending message to the front-end]
await self.send(data)
\end{lstlisting}

Creating the new path can be done adding a new path to ws\_urlpatters list:

\begin{lstlisting}[language=Python, caption=Mapping path to the new Consumer]
ws_urlpatterns = [
  path('ws/connection/', ConnectionConsumer.as_asgi()),
  path('ws/receive/', ReceiveCommandConsumer.as_asgi()),
  path('ws/new-socket/', NewConsumer.as_asgi()),
]
\end{lstlisting}

To start a websocket connection, from the Front-end, it is necessary to create a new object on the JavaScript file, passing as argument an available URL in the routing schema.

\begin{lstlisting}[language=JavaScript, caption=New websocket connection request]
var socket = new WebSocket('ws://localhost:8000/ws/new-socket/');
\end{lstlisting}

Finnaly, accessing ws://localhost:8000/ws/new-socket/, a dedicated full-duplex connection should be stablished and our two ends can communicate with each other.

This WebSocket object created has methods to interact with the socket connection. Here are the main ones:

\begin{itemize}
    \item send: Transmits data to the server via the WebSocket connection
\begin{lstlisting}[language=JavaScript, caption=Send a JSON to the back-end]
socket.send(jsonToSend);
\end{lstlisting}
    \item readyState: The current state of the connection, this is one of the Ready state constants. Read-only.
\begin{lstlisting}[language=JavaScript, caption=Use readyState to check connection status]
if (socket.readyState == WebSocket.OPEN) {
    // Handle connection OPEN
}
\end{lstlisting}
    \item onclose: An event listener to be called when the readyState of the WebSocket connection changes to CLOSED.
\begin{lstlisting}[language=JavaScript, caption=Use onclose to handle disconnection]
socket.onclose = function(e) {
  // Handle connection closed
};
\end{lstlisting}
    \item onmessage: An event listener to be called when a message is received from the server. Receives a message parameter.
\begin{lstlisting}[language=JavaScript, caption=Use onmessage to handle receiving a message]
socket.onmessage = function(msg) {
  // Handle message received
}
\end{lstlisting}
\end{itemize}

\subsubsection{External Communication}

The primary purpose of this framework is to exchange information with other devices. Currently, there are implemented two ways for external connections.

The first way is to plugin an ESP32 microcontroller to the framework's machine. This microcontroller should be able to detect other devices that receive and send information to them. Our framework can establish a UART connection with a plugged ESP32 microcontroller, receive everything sent via serial, and send commands via serial, making the microcontroller responsible for retransmitting the command. In order to accept a connection with an ESP32 microcontroller, it is necessary to insert the correct UART Port and baud rate inside \verb|config.ini|. The \verb|SerialConnection| class, from \verb|/connections/serial\_connector.py|, is instantiated when javascript starts a WebSocket connection of this type. The instantiated object keeps trying connection with the UART Port. Once a microcontroller is plugged in, the interface indicates this change, and you can exchange information through the ESP32 microcontroller.

Another way to communicate with our framework is with POST requests. A device, an UAV (drone) per se, wants to send its location to our ground station. This can be achieved with a POST request to the specific ground station URL.

\begin{lstlisting}[language=Python, caption=UAV code to send it's information to the ground station]
json_tmp = {"id": uav_id, "lat": targetpos.lat, "lng": targetpos.lng, "alt": targetpos.alt, "ip": args.uav_ip + ':' + flask_port}

r = requests.post(path_to_post, data=json_tmp)
\end{lstlisting}

Note that the device should attach, on the message, it's own IP and PORT, so our framework can send commands back to it. The specific URL, to receive POSTs, is mapped to a view. So, when the device send it's location on body's request, the post\_to\_socket view receive the request and extracts the information from it's body. We want to send this information to our interface and also to save it in the log file. Who is responsible for both actions is the \verb|PostConsumer|, inside \verb|/connections/consumers_wrapper/post_consumer.py|. This way, the post\_to\_socket view needs to send the message to \verb|PostConsumer|, getting an instance of this class and calling this Class function receive\_post(message).

\begin{lstlisting}[language=Python, caption=View receiving external info and sending to PostConsumer]
post_consumer_instance = get_post_consumer_instance()
await post_consumer_instance.receive_post(new_dict)
\end{lstlisting}

Sending a message to an external device is also done by Consumers. When a command button is activated on the interface, the \verb|main.js| uses the async method socket.send(), to transmit the command direct to the Consumer (back-end). The message received from the \verb|main.js|, contains which device or group of devices it should be sent. It also contains the ID of the external devices that will receive the command. The first step is to search on the registered device's list for the address (IP) of the devices.

\begin{lstlisting}[language=Python, caption=Getting the IP based on the ID]
device_to_send_list = get_device_from_list_by_id(device_receiver_id)
\end{lstlisting}

There is a list on \verb|config.ini| mapping the commands code (integer) to a specific endpoint, that should be added to the IP+Port of the external device.

\begin{lstlisting}[language=Python, caption=List of commands]
[commands-list]
20 = position_absolute_json,get
22 = position_relative_json,get
24 = auto,get
26 = run_experiment,get
28 = set_auto,get
30 = rtl,get
32 = takeoff_and_hold,get
\end{lstlisting}

The list contains the endpoint and the HTTP request type, if it is a GET or POST request.

With the address complete, the command will be sent via HTTP request.

\begin{lstlisting}[language=Python, caption=Sending a HTTP request containing the command to the specific device]
command_path_list = config['commands-list'][command].split(',')
endpoint = command_path_list[0]

if command_path_list[1] == 'get':
  #GET request
  task = asyncio.create_task(self.send_get_specific_device(url, id, device['device']))
else:
  # POST request
  task = asyncio.create_task(self.send_post_specific_device(url, json_to_send))
self.async_tasks.append(task)
\end{lstlisting}

Depending on the type of the request, the command will be sent and an asynchronous task will be created.

\subsection{Data persistence}
One of the main features of this project is the data persistence of every event that occurred during the experiments. Log files are generated, when starting the application, and filled in as messages are received, errors are caught, commands are sent, and other events that are of importance to the experiment.

To generate the \verb|.log| files, the logging package, for Python, is used. Inside \verb|/connections/utils/logger.py| there is a class Logger, responsible for the persistent logic. It's possible to extend and copy this class to other modules, for example, at the \verb|uav_simulator/| module that has this class with different logic.

When the server start, a .log file is created, inside the folder specified by the Logger's \verb|path| variable, and the file's name is composed by the module name followed by the date created. The example above represents a .log file created inside the uav\_simulator module at 25/01/2022 08:18:40.

\begin{lstlisting}[language=Python, caption=Log file name]
uav_simulator-2022-01-25-08-18-40.log
\end{lstlisting}

To fill this file, it must be inserted in code calls of the methods from the Logger class, according to it's needs. The example above includes the code from the PostConsumer class, inside the method to handle a external message received.

\begin{lstlisting}[language=Python, caption=Calling the two different methods from Logger class]
logger.log_info(source=source, data=data, code_origin='receive-info')
try:
  await self.send(json.dumps(data)) # Send to JS via socket
except Exception:
  logger.log_except()
\end{lstlisting}

The logger object is global and already instantiated. Two log methods are called, to save the data received and to save the Exception caught when trying to send the message to the front-end via socket.

The .log file format is specified inside the Logger class, using the syntax accepted by the Formatting class, form logging package. For more information on how to format the .log file, https://docs.python.org/3/library/logging.html\#logging.Formatter.

\begin{lstlisting}[language=Python, caption=The body of a .log file]
2021-12-12 20:58:32,706; uav-21; receive-info; {'id': 21, 'type': 102, 'seq': 30, 'lat': -15.840081, 'lng': -47.926642, 'alt': -0.03, 'device': 'uav', 'ip': 'http://127.0.0.1:5071/', 'method': 'post', 'time': '2021-12-12T20:58:32.706792', 'status': 'active'}

2022-01-25 20:58:50,365; gs; send-get; http://127.0.0.1:5071/rtl
\end{lstlisting}

The example above has two messages, formatted with the date of the event, who triggered the event, where it was triggered and the message itself.

\section{Usage}

\subsection{Installation}

\textbf{Prerequisites}

In order to use the components in this repository, you need to have Python 3.0 or higher installed. Also, pip, a Python package manager, is recommended to manage and automatically install the required packages of this project. After installing Python, pip should be installed by default. You can check if it's already installed and its version:

\begin{lstlisting}[language=Python, caption=Command line to check if pip is installed]
C:\> python -m pip --version
\end{lstlisting}

\textbf{Cloning the repository}

With Python3 installed, you should be able to clone the repository: https://github.com/BrenoFischer/gradys-gs

\textbf{Creating a virtual environment}

In order to keep this framework in a separate environment, with it's own packages and versions, it's recommended to create a virtual environment. On Windows:

\begin{lstlisting}[language=Python, caption=Command line to create a virtual environment]
Windows
C:\> python -m venv C:\path-to-this-cloned-repository/venv
\end{lstlisting}

On Linux, you can check if virtualenv is already installed, install it, if not already installed, and create the venv:

\begin{lstlisting}[language=Python, caption=On Linux how to check venv install and create a new venv]
Linux
gradys-gs$ virtualenv --version
  virtualenv xx.x.x
gradys-gs$ sudo pip3 install virtualenv
gradys-gs$ virtualenv venv
\end{lstlisting}

This will create a folder called venv, inside the project's folder. Now you have to activate the environment to install/use packages only from this venv.

\begin{lstlisting}[language=Python, caption=Activating the virtual environment]
Windows
C:\> C:\path-to-this-cloned-repository\venv\Scripts\activate
\end{lstlisting}

\begin{lstlisting}[language=Python, caption=Activating on Linux the virtual environment]
Linux
gradys-gs$ source venv/bin/activate
\end{lstlisting}

\textbf{Installing necessary packages}

The list of necessary packages are inside requeriments.txt file, if you are using Windows. It'll be installed automatically, using the Python package manager, pip. You can install, running on Windows console:

\begin{lstlisting}[language=Python, caption=Using requeriments.txt and pip to install the necessary packages]
Windows
C:\path-to-this-cloned-repository\> pip install -r requeriments.txt
\end{lstlisting}

On Linux, you should run the compatible script file, requeriments\_linux.txt:

\begin{lstlisting}[language=Python, caption=Installing the necessary packages on Linux]
Linux
gradys-gs$ pip install -r requeriments_linux.txt
\end{lstlisting}

\textbf{Secret variables}

This project uses Google Maps services, with paid features. To use these functionalities you need to have or create a Google Maps API Key. Google's guide on how to create an API Key: https://developers.google.com/maps/gmp-get-started
This project also use Django Framework that has a secret key variable, for security purposes. You can generate your Django secret key here: https://djecrety.ir/

The framework will load automatically these as environment variables. With both private keys created,

\begin{itemize}
    \item Inside /config/.env, insert the secret keys:
        \begin{itemize}
            \item SECRET\_KEY='xxxx' Changing 'xxxx' with your Django secret key
            \item GOOGLE\_MAPS\_API\_KEY='xxxx' Changing 'xxxx' with your Google Maps key
        \end{itemize}
\end{itemize}

\textbf{Running the server}

Django provides lightweight development Web server, that you can use via manage.py file. By default, the server runs on port 8000 on the IP address 127.0.0.1 and should not be used on production. You can run with:

\begin{lstlisting}[language=Python, caption=Starting the server]
Windows
C:\path-to-this-cloned-repository\> python manage.py runserver
\end{lstlisting}

\begin{lstlisting}[language=Python, caption=Starting the server on Linux]
Linux
gradys-gs$ python manage.py runserver
\end{lstlisting}

Or, with diferent IP/PORT, in the example below, Port 8000 on IP address 0.0.0.0. This IP is will listen to all IP adresses the machine supports. So for example, with this server configuration up, you can open the web navigator with localhost:8000 and the inet ip obtainable from ifconfig (linux environment):

\begin{lstlisting}[language=Python, caption=Starting the server accepting all IP addresses the machine supports]
C:\path-to-this-cloned-repository\> python manage.py runserver 0.0.0.0:8000
\end{lstlisting}

Remember to insert, inside config.ini file, the correct IP + Port, on [post] category, if changed to a specific IP, when running the command above.

\textbf{Connecting to home page}

Now you should be able to connect to the home page, acessing, on your browser, the IP/PORT the server is up, on default: localhost:8000.

\subsection{Extension}
Since the project is Open Source anyone is able to download the source code, experiment and create new things as the project's necessities. The project's architecture was organized with extension in mind and it especially facilitates the development of new form of external communications, sets of buttons functionalities and logging formats.

\subsubsection{Interface composition}
The interface is separated into four sections:

\begin{itemize}
    \item Virtual map
    \item Command buttons
    \item Text log
    \item Connection status
\end{itemize}

\begin{figure}[h]
    \centering
    \includegraphics{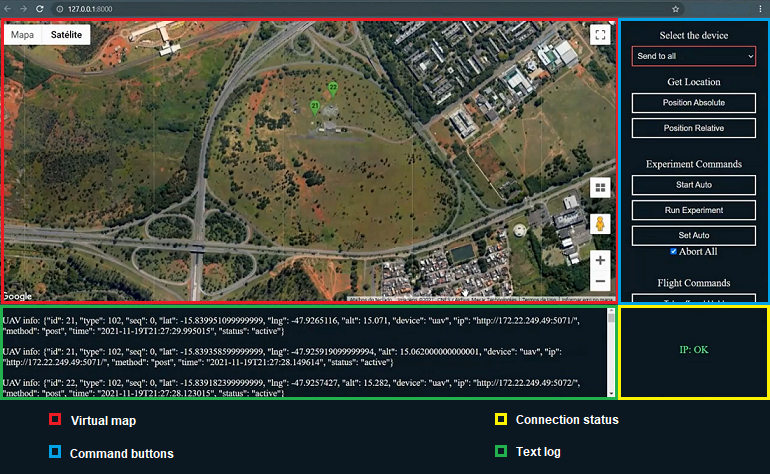}
    \caption{Interface composition}
\end{figure}

\textbf{Virtual map section}: Google's virtual map occupies the biggest section. It's an interactive map, scrollable and capable of zoom in/zoom out. Note that an internet connection is needed to load the map, after the web page is rendered. Javascript files handle the virtual map's logic, automatically updating the devices location and status, upon receiving JSON messages from the back-end.

Upon receiving an information from the back-end, the Javascript method above will be called, updating the map.

\begin{lstlisting}[language=JavaScript, caption=Function to update the virtual map markers]
//Search if the device is already on map
let foundedMarkerIndex = this.findMarkerIdIndex(id);

    if(foundedMarkerIndex == -1) { //Not on map, create a new marker
      const image = this.getMarkerImage(id, status, deviceType);

      let myLatLng = new google.maps.LatLng(lat,lng);
      let marker = new google.maps.Marker({
        position: myLatLng,
        icon: image,
      });
      //Insert the new mark on devices list
      this.markers.push(new MyMarker(id, marker));
      //Render the marker on map
      marker.setMap(this.map);
    }
    else { //Already on map, update marker
      const myLatLng = new google.maps.LatLng(lat,lng);
      const image = this.getMarkerImage(id, status, deviceType);

      this.markers[foundedMarkerIndex].marker.setIcon(image);
      this.markers[foundedMarkerIndex].marker.setPosition(myLatLng);
    }
\end{lstlisting}

\textbf{Command buttons section}: the command buttons section contains a select field, to select a group or a specific device. All commands sent will be only to the selected option. This section also contains subsections of buttons, with a title and the buttons set. A button can be a checkbox button, that'll trigger the command while checked and a single action button, that'll trigger an action on click. More buttons can be created, explained in detail in the subsection "Creating new buttons".

\textbf{Text log section}: the text log section shows all messages received by the ground station, in white text, and sent by the ground station, in red text.

\textbf{Connection status section}: the connection status section shows if the multiple connections are available to use.

\subsubsection{Creating new buttons}

It's possible to register a new button inside the template, create a onClick callback function and send the command via websocket to Django (back-end).

\begin{itemize}
    \item Create new button in /templates/index.html
\begin{lstlisting}[language=Python, caption=New button on /templates/index.html]
<input class="button" id="new-button" type="button" value="New Command">
\end{lstlisting}
    \item Register an onclick function, in /static/connections/main.js.
\begin{lstlisting}[language=JavaScript, caption=Register onclick and code number on /static/connections/main.js]
var newCommandNumber = 40

document.querySelector('#new-button').onclick = function(e) {
  sendCommand(newCommandNumber);
};
\end{lstlisting}
    \item Send to the back-end, when button is clicked
\begin{lstlisting}[language=JavaScript, caption=Button's logic on /static/connections/main.js]
function sendCommand(type) {
  jsonToSend = {id: 1, type: type}
  // ...
  if (socket.readyState == WebSocket.OPEN) {
    socket.send(jsonToSend);
  }
}
\end{lstlisting}
\end{itemize}

Notice that the socket object must be instatiated already, and the connection 'OPEN'. The corresponding Consumer will receive the message and handle, acording to it's command type.

You already have a button on interface that sends a command, in this case '40', to a Consumer. This Consumer will be in charge to the command logic.
Inside the 'receive' method of this Consumer's Class, it's up to you to write the command's logic, according to your communication protocol.
When handling with HTTP requests, you can insert the new command to the command's list, inside config.ini file. The Consumer can iterate this list and check the command received, mapping to the right endpoint.

\begin{lstlisting}[language=Python, caption=Commands list on config.ini]
[commands-list]
20 = position_absolute_json,get
22 = position_relative_json,post
32 = takeoff_and_hold,get
...
\end{lstlisting}

This list contains a number as the key, to the corresponding endpoint address, that will receive the HTTP request. The type of request is represented after the comma, with no spaces. If your communication is using HTTP requests and this list, your new list, with the new command, should look like this:

\begin{lstlisting}[language=Python, caption=New command on list]
[commands-list]
20 = position_absolute_json,get
22 = position_relative_json,post
32 = takeoff_and_hold,get
...
40 = new_endpoint,get
\end{lstlisting}

\subsubsection{Changing parameters}
Some of the framework's informations are initialized by the \verb|config.ini| file. This file is located on the project root folder. Below are listed the parameters that can be changed.

\textbf{UART Connection}

One way this framework can comunicate with a network is with a dedicated ESP32, using UART Protocol. The ESP device connected via serial has a specific PORT and Baudrate, that can be changed inside config.ini with the [serial] tag:

\begin{lstlisting}[language=Python, caption=Serial connection parameters on config.ini]
[serial]
# The serial PORT the ESP32 is connected
port = COM4

# Rate of information transferred in the serial port
# Needs to be the same in ESP32 connection
baudrate = 115200

# If this Protocol is used
serial_available = false
\end{lstlisting}

\textbf{POST request}

Another way to comunicate with nodes of the network is receiving/sending information via POST/GET Requests. Django provides a routing system that acessibles URLs trigger methods, or Views. A device can send a POST request to http://127.0.0.1:8000/update-info/ (or IP/PORT running the application). Notice that a device should send inside the message it's own IP/PORT, so the application can send commands via HTTP requests. This structure is described with more details below, on the Project Struct topic.

Inside the config.ini file, below the [post] tag, you can change some of the protocol's variables:

\begin{lstlisting}[language=Python, caption=Post Request parameters on config.ini]
[post]
# Default ip/port of django's server.
# If started on a different configuration, you need to change it here.
ip = http://127.0.0.1:8000/

# Endpoint that'll receive POST requests with device's information
path_receive_info = update-info/
\end{lstlisting}

The application saves the latest messages of unique devices in a list, inside the update\_periodically\_consumer.py, for each execution. From time to time, it's sent to the front-end, via web-socket, with the activity status of each device. A device can be active, on hold and inactive, depending on the interval of it's last message. These variables can be adjusted in the config.ini file, below the [list-updater] tag:

\begin{lstlisting}[language=Python, caption=Device List Updater parameters on config.ini]
[post]
# The amount of seconds to a device be considered 'inactive'
seconds_to_device_be_inactive = 50

# The amount of seconds to a device be considered 'on hold'
seconds_to_device_be_on_hold = 25

# The amount of seconds to update the list of devices in Front-End
update_delay = 20
\end{lstlisting}

\subsubsection{Inserting new external communication protocols}

This framework is designed to accept different forms of external communication, providing the internal structure to assist the experimental tests. The logic and code of the new communication is up to the developer to write and this section will guide on this process.

Because the communication module of the ground station is built using Django, the new protocol code have to be written using the Python Programming Language. A new Consumer can be created on \verb|/connections/consumers_wrappers/|, to handle the internal messages, exchanging via WebSocket with the Front-end. How the message will reach this Consumer is up to the developer. An example is cited on the External Communication subsection, where a View receive the external information and instantiate a Consumer object, to forward the message received.

With the Consumer created and connected with the new external communication module, a new WebSocket connection should be created, as explained on Internal Communication subsection.

The way the Consumer will handle commands from the interface should be written on \verb|receive(data)| method.

\section{Conclusion}

This work is a step within a set of deliverables for a project. The GrADyS project uses this tool to validate previously simulated protocols and compare field test results with accurate sensors and UAVs.

At the moment, GrADyS-GS can connect and control swarms of UAVs and WSN by organizing and facilitating field experiments that are previously simulated on GrADyS-SIM\cite{gradyssim2022} which new Bluetooth routing algorithms such as MAM\cite{paulon2022}.

This tool is in whole evolution and is being used in real tests, which facilitates and increases its functionality in a natural, practical, and result-oriented way.

In our roadmap, there is the integration with new devices and a Bind with the discrete event simulator OMNET++/INET.

We invite the open source community to fork it, use it, and contribute at https://github.com/BrenoFischer/gradys-gs. 

\section*{Acknowledgments}
This study was financed in part by AFOSR grant FA9550-20-1-0285.

\bibliographystyle{unsrt}  
\bibliography{references}

\begin{thebibliography}{1}

\bibitem{gradys2021}
Bruno~Jos{\'{e}} {Olivieri De Souza}, Marcelo Paulon, Juc{\'{a}} Vasconcelos,
  and Markus Endler.
\newblock {GrADyS: Exploring movement awareness for efficient routing in
  Ground-and-Air Dynamic Sensor Networks}.
\newblock dec 2020.

\bibitem{bolivieri2020}
Bruno~Jos{\'{e}} {Olivieri de Souza} and Markus Endler.
\newblock {Evaluating flight coordination approaches of UAV squads for WSN data
  collection enhancing the internet range on WSN data collection}.
\newblock {\em Journal of Internet Services and Applications}, 11(1):4, dec
  2020.

\bibitem{gradyssim2022}
Thiago Lamenza, Marcelo Paulon, Breno Perricone, Bruno~José Olivieri~de Souza,
  and Markus Endler.
\newblock {GrADyS-SIM - A OMNET++/INET simulation framework for Internet of
  Flying things}.
\newblock {\em arXiv.org}, pages 1--9, 2022.

\bibitem{paulon2022}
Marcelo {Paulon J.V.}, Bruno~Jos{\'{e}} {Olivieri de Souza}, and Markus Endler.
\newblock {Exploring data collection on Bluetooth Mesh networks}.
\newblock {\em Ad Hoc Networks}, 130(February):102809, 2022.

\end{thebibliography}

\end{document}